\documentclass[letterpaper]{article}
\usepackage{aaai17}
\usepackage{times}
\usepackage{helvet}
\usepackage{courier}

\usepackage{graphicx}
\usepackage{latexsym}
\usepackage{mathptmx}
\usepackage{amsmath,amssymb}
\usepackage{dsfont}
\usepackage{theorem}

\newcommand{\defined}{\stackrel{\rm def}{=}}

\newcommand{\mbf}[1]{ {\mathbf #1} }

\newcommand{\denote}[1]{ [\![ #1 ]\!] }

\begin{document}

\title{Embedding Tarskian Semantics in Vector Spaces}


\author{Taisuke Sato \\
AI research center AIST / 
National Institute of Informatics, Japan
 \\
email:satou.taisuke@aist.go.jp
}

\maketitle


\begin{abstract}
We  propose a  new linear  algebraic  approach to  the computation  of
Tarskian semantics  in logic.  We  embed a  finite model ${\bf  M}$ in
first-order logic with $N$ entities in $N$-dimensional Euclidean space
$\mathds{R}^N$ by  mapping entities  of ${\bf  M}$ to  $N$ dimensional
one-hot vectors  and $k$-ary relations to  order-$k$ adjacency tensors
(multi-way  arrays).  Second  given a  logical formula  $F$ in  prenex
normal  form,  we compile  $F$  into  a  set $\Sigma_F$  of  algebraic
formulas in multi-linear algebra with  a nonlinear operation.  In this
compilation, existential quantifiers are compiled into a specific type
of tensors,  e.g., identity  matrices in the  case of  quantifying two
occurrences of a  variable.  It is shown that  a systematic evaluation
of  $\Sigma_F$ in  $\mathds{R}^N$ gives  the truth  value, 1(true)  or
0(false),  of $F$  in ${\bf  M}$.  Based  on this  framework, we  also
propose an unprecedented way of  computing the least models defined by
Datalog programs in linear spaces via matrix equations and empirically
show its effectiveness compared to state-of-the-art approaches.
\end{abstract}

\section{Introduction}

In  this paper,  we propose  a new  linear algebraic  approach to  the
computation of  Tarskian semantics,  i.e., the standard  semantics for
first-order  logic.  Tarskian  semantics  determines  the truth  value
$\denote{F}$ of first-order formulas $F$ in a model ${\bf M}$ based on
a relational structure comprised of  a non-empty domain ${\cal D}$ and
relations  over ${\cal  D}$, using  an interpretation  associated with
${\bf M}$ that maps constants to  entities in ${\cal D}$ and predicate
symbols to the relations.   $\denote{F}$ is step-by-step determined in
${\bf M}$ along the syntactic structure  of $F$.  What we propose here
is to carry out this evaluation in another model isomorphically copied
to  the  $N$-dimensional  Euclidean  space  $\mathds{R}^N$,  when  the
first-order  language ${\cal  L}$ we  use has  only $N$  constants and
correspondingly ${\cal D}$ contains $N$ entities.

More  precisely, given  a  finite  model ${\bf  M}$,  we first  encode
entities in ${\cal D}$ into vectors in $\mathds{R}^N$ where $N$ is the
cardinality of ${\cal  D}$ and also encode $k$-ary  relations in ${\bf
  M}$ to order-$k$ adjacency tensors in multi-linear algebra.  Then to
evaluate a  logical formula $F$  in prenex normal form,  starting from
atoms, we inductively derive a set $\Sigma_F$ of algebraic formulas in
multi-linear algebra augmented with a nonlinear operation.  Evaluating
$\Sigma_F$  in $\mathds{R}^N$  gives the  truth value  $\denote{F}$ in
{\bf  M},  that is,  $\denote{F}  =1$  if  ${\bf  M} \models  F$  else
$\denote{F}=0$.\\

Our  proposal  is  motivated  by  recent  work  on  logical  inference
concerning knowledge graphs(KGs).  KGs are graphs encoding RDF triples
of the form $({\rm subject}:s,\, {\rm predicate}:p,\, {\rm object}:o)$
and can be considered  as a set of ground atoms  of the form $p(s,o)$.
There are  huge  KGs available such  as Freebase\cite{Bollacker08}.
%
%
%
The problem is that although they are good resources of the real world
information and  logically simple, they  are huge, containing  tens of
millions  of  nodes and  furthermore  incomplete;  there are  lots  of
inconsistent data and also lots of missing data.  To carry out various
KG  tasks  such  as  computing  the truth  value  (or  more  generally
probability) of $p(s,o)$  while coping with the sheer  amount of data
and   incompleteness,    three   major   approaches    are   developed
\cite{Nickel15}; one that  is based on probabilistic  models, one that
uses  explicit features  sampled from  the graph  and one  that learns
latent  feature vectors  from the  graph.  The  last approach,  latent
feature approach, compiles entities and  predicates in the domain into
vectors  and  tensors\cite{Kolda09}  respectively  and  apply  various
linear algebraic operations, with  dimension reduction, to compute the
probability of ${p(s,o)}$.

In the development  of these approaches, formulas  beyond ground atoms
are  introduced  and  investigated such  as  existentially  quantified
conjunctions as  queries and  definite clauses  as constraints  on KGs
\cite{Grefenstette13,Rocktaschel15b,Krompass14,Guu15,Yang15}.
However, from a logical point of view, their treatment was confined to
propositional logic  level and  the evaluation of  general first-order
formulas is  left untouched except  for the work done  by Grefenstette
\cite{Grefenstette13}.  Regrettably, while  he succeeded in completely
embedding   the   fragment   of   model  theory,   model   theory   of
quantifier-free  first-order  logic,   in  tensor  spaces,  quantified
formulas were  excluded and  had to be  treated separately  by another
framework.  Nested  quantification was not  allowed either. So  how to
evaluate  arbitrarily  quantified formulas  in  a  vector space  still
remains open.

We solve this problem by  introducing specific tensors for existential
quantifiers together with a  nonlinear operation.  Our contribution is
two-fold. First we introduce a  single framework for the evaluation of
quantified first-order formulas in a vector space, assuming the domain
is finite, thus solving the remaining problem.

The second contribution is to present  a concrete method, based on our
framework, to compute  the least model of Datalog programs  in a vector
space, which  opens up  a completely new  way of  evaluating recursive
programs, though we  have to skip details due to  page limitations and
only sketch experimental result.

%

At this  point it would  be beneficial  to ask why  evaluating logical
formulas in a vector space is an interesting idea.  First, there are a
rich family of  algebraic operations available in a  vector space such
as inner product,  outer product, projection, PCA, SVD and  so on that
helps analyzing and manipulating vector data.   Second, basically they
are  of  polynomial time  complexities,  so  we can  expect  efficient
computation.  Last but not least, approximation through various matrix
and tensor  decomposition potentially  leads to logical  inference for
Web scale symbolic data.

We  assume the  reader is  familiar with  basics of  logic and  linear
algebra including  matrices and  tensors\cite{Kolda09,Cichocki09}.

\section{Preliminaries}
We first review  some terminology in logic. We assume  our first order
language ${\cal  L}$ contains $N$ constants  $\{e_1,\ldots,e_N \}$ and
no function symbols.

A model $\mathbf  M=({\cal D},I)$ is a pair of  domain, a nonempty set
${\cal D}$  and an interpretation  $I$ that maps constants  ${e_i}$ to
elements  (entities,  individuals) $I(e_i)  \in  {\cal  D}$ and  k-ary
predicate  symbols  $r$  to  k-ary  relations  $I(r)  \subseteq  {\cal
  D}\overbrace{\times\cdots\times}^{k\;{\rm   times}}{\cal  D}$.    An
assignment $a$ is a mapping from variables $x$ to an element $a(x) \in
{\cal D}$.  It  provides a way of evaluating  formulas containing free
variables.  Syntactically  terms mean  variables and/or  constants and
atomic  formulas  or  atoms  $r(t_1,\ldots,t_k)$ are  comprised  of  a
$k$-ary predicate  symbol $r$ and  $k$ terms $t_1,\ldots,t_k$  some of
which may  be variables.  Formulas  $F$ in ${\cal L}$  are inductively
constructed as  usual from  atoms using logical  connectives (negation
$\neg$,  conjunction  $\wedge$,  disjunction $\vee$)  and  quantifiers
($\exists$,$\forall$).

Now we define free/bound occurrences of variables in $F$.  When $F$ is
an  atom, all  variables in  $F$ occur  free in  $F$.  When  $F$ is  a
negation $\neg  F_1$, disjunction $F_1  \vee F_2$ or  conjunction $F_1
\wedge F_2$,  free variables in  $F_1$ and  those in $F_2$  both occur
free in $F$  and vice versa.  When $F$ is  an existentially quantified
formula $\exists x F_1$, free variables in $F_1$ except $x$ occur free
in $F$ and vice versa.  Variables in $F$ that do not occur free in $F$
are said to be bound.  A formula  is closed if it has no free variable
whereas it is open if it has no quantification.

Given a  model $\mathbf  M=({\cal D},I)$ and  an assignment  $a$, the
denotation $\denote{E}_{I,a}$ in $\mathbf{M}$  of an expression $E$ is
inductively  defined  for  terms  $t$ and  formulas  $F$  as  follows.
$\denote{t}_{I,a} =I(t)  \;\mbox{if $t$ is a  constant, else}\; a(t)$.
When      $r$       is      a      $k$-ary       relation      symbol,
$\denote{r(t_1,\ldots,t_k)}_{I,a}     =     1({\rm    true})$     {if}
$(\denote{t_1}_{I,a},\ldots,\denote{t_k}_{I,a})   \in    I(r)$,   else
0(false).  Let $F_1,F_2$ be formulas.  For a negation ${\neg F_1}$, we
define    $\denote{\neg   F_1}_{I,a}    =1   -    \denote{F_1}_{I,a}$.
$\denote{F_1   \vee  F_2}_{I,a}   =1$  if   $\denote{F_1}_{I,a}=1$  or
$\denote{F_2}_{I,a}=1$, else 0.  $\denote{\exists  x F_1}_{I,a} =1$ if
there    exists     some    $d    \in    {\cal     D}$    such    that
$\denote{F_1}_{I,a[x\leftarrow  d]}=1$, else  0.  Here  $a[x\leftarrow
  d]$ is  a new  assignment that  is the  same as  $a$ except  that it
assigns $d$ to the variable $x$.  Finally $\denote{A \wedge B}_{I,a} =
\denote{\neg  (\neg  A \vee  \neg  B)}_{I,a}$  and $\denote{\forall  x
  F_1}_{I,a}  =  \denote{\neg \exists  x  \neg  F_1}_{I,a}$.  For  any
formula $F$, $\denote{F}_{I,a} \in \{1,0\}$ and when $\denote{F}_{I,a}
= 1$, we write ${\mathbf M} \models_a F$.  However when $F$ is closed,
since $\denote{F}_{I,a}$  does not  depend on  the assignment  $a$, we
just write $\denote{F}$ and ${\mathbf M}  \models F$ if $F$ is true in
$\mathbf M$.  For formulas $F,G$, we say $F$ and $G$ are equivalent and
write $F  \equiv G$ if  $\denote{F}_{I,a} = \denote{G}_{I,a}$  for any
model ${\mathbf M}$,  any interpretation $I$ and  any assignment $a$.
In  what follows,  $I,a$  are omitted  when they  are  clear from  the
context.

Then  recall that  a  literal is  an atom  (positive  literal) or  its
negation (negative literal).  Suppose $F$ is an open formula.  $F$ has
an   equivalent    formula   in   disjunctive   normal    form   (DNF)
$A_1\vee\cdots\vee A_k$ such  that each disjunct $A_i$  is a monomial,
i.e., conjunction of  literals.  Dually $F$ has  an equivalent formula
in conjunctive normal form  (CNF) $A_1\wedge\cdots\wedge A_k$ such that
each conjunct $A_i$ is a clause, i.e., disjunction of literals.

It is  known that every  formula has  an equivalent formula  in prenex
normal  form  $Q_1x_1\cdots  Q_mx_m   G$  where  $Q_1,\ldots,Q_m$  are
quantifiers $\exists,  \forall$ and $G$  is open.  So to  evaluate the
truth value $\denote{F}$  of a given $F$ in $\mathbf  M$, since $G$ is
equivalent to  DNF or CNF,  we have only  to evaluate a  prenex normal
form $Q_1x_1\cdots Q_mx_m G$ where $G$ is an open DNF or CNF.

Note the subformula $Q_mx_m G$.  Since  DNF and CNF are convertible to
each other, it  is equivalent to $\exists$-DNF  or $\forall$-CNF where
$\exists$-DNF  is  a  class  of  formulas  $F$  such  that  $F$  is  a
disjunction   of   disjuncts  $\exists\,x(L_1\wedge\dots\wedge   L_M)$
comprised  of literals  $L_1,\ldots,L_M$, whereas  $\forall$-CNF is  a
class of  formulas $F$  such that  $F$ is  a conjunction  of conjuncts
$\forall\,x(L_1\vee\dots\vee L_M)$.  Suppose  $F = Q_1x_1\cdots Q_mx_m
G$ is given. We may assume, without loss of generality, that if $Q_m =
\exists$, $Q_mx_m G$ is in  $\exists$-DNF.  Otherwise $Q_mx_m G$ is in
$\forall$-CNF.

Now we  turn to  vector spaces.  We  consider tensors  as multi-linear
maps as  mathematical objects and  multi-way arrays as  data structure
depending on  the context\cite{Kolda09,Cichocki09}.   Although tensors
are  a generalization  of vectors  and matrices,  we specifically  say
vectors and matrices when their shape needs to be distinguished.

In what follows,  scalars are denoted by lower case  letters like $a$.
Vectors mean column vectors and we  denote them by boldface lower case
letters like  $\mbf{a}$ and  $\mbf{a}$'s components by  $a_i$.  ${\cal
  D}'= \{ \mathbf{e}_1,\ldots,\mathbf{e}_N \}$ stands for the standard
basis   of  $N$-dimensional   Euclidean  space   $\mathds{R}^N$  where
$\mathbf{e}_i =  (0\cdots,1,\cdots,0)^T$ is a  vector that has  one at
the  $i$-th position  and zeros  elsewhere.  Such  vectors are  called
one-hot   vectors.    $\mathbf{1}$   is   a  vector   of   all   ones.
$(\mathbf{a}\bullet \mathbf{b}) = \mathbf{a}^T\mathbf{b}$ is the inner
product  of $\mathbf{a}$  and $\mathbf{b}$  whereas $\mathbf{a}  \circ
\mathbf{b} = \mathbf{a}\mathbf{b}^T$ is their outer product.
Matrices are assumed  to be square and written by  boldface upper case
letters like  $\mbf{A}$.  In  particular $\mbf{I}$ is  an identity
matrix. $\mathds{1} =  \mbf{1}\circ \mbf{1}$ is a  matrix of all
ones. ${\rm tr}(\mbf{A})$ stands for the trace of $\mbf{A}$.
Order-$p$  tensors $\in \mathds{R}^{\overbrace{N\times\cdots\times N}^p}$
are   denoted  by   underlined  matrices   like
${\underline{\mbf  A}}$ or $\{a_{i_1,\ldots,i_p}\}$
($1 \leq i_1,\ldots,i_p \leq N$).
${\underline{\mbf  A}}$'s component $a_{i_1,\ldots,i_p}$
is also written as $(\underline{\mbf  A})_{i_1,\ldots,i_p}$.
Let $\underline{\mbf A}=\{ a_{i_1,\ldots,i_p} \}$ and
$\underline{\mbf B}=\{ b_{k_1,\ldots,k_q} \}$ be tensors.
The mode-($n$,$m$) contracted product
$\underline{\mbf A} \times_{n,m} \underline{\mbf B}$ of
$\underline{\mbf A}$ and $\underline{\mbf B}$
is defined by
$(\underline{\mbf A} \times_{n,m}
\underline{\mbf B})_{i_1,\dots,i_{n-1},i_{n+1},\ldots,i_p,k_1,\ldots,k_{m-1},k_{m+1},\ldots,k_q}
$
$=\sum_{j} a_{i_1,\dots,i_{n-1},j,\ldots,i_p}b_{k_1,\ldots,k_{m-1},j,\ldots,k_q}$
with the convention that the association is to the left, i.e.,
$\underline{\mbf A} \times_{n,m} \underline{\mbf B}
\times_{p,q} \underline{\mbf C} =
(\underline{\mbf A} \times_{n,m} \underline{\mbf B})
\times_{p,q} \underline{\mbf C}$.
So $\underline{\mbf  A}  \bullet_n   {\mbf  u}$,
the contracted product of $\underline{\mbf  A}$ and vector
${\mbf  u}$, which is computed by
$(\underline{\mbf A} \bullet_n {\mbf u})_{i_1,\dots,i_{n-1},i_{n+1},\ldots,i_p}
= \sum_{j} a_{i_1,\dots,i_{n-1},j,\ldots,i_p}u_{j}$
is equal to $\underline{\mbf  A} \times_{n,1} {\mbf  u}$
and the usual $n$-mode  product
${\underline  {\mbf  A}}  \times_n   {\mbf  U}$
of $\underline  {\mbf  A}$ and matrix ${\mbf  U}$
is equal to $\underline{\mbf  A}   \times_{n,2}  {\mbf  U}$.

Tensors  can be  constructed by outer products;
$(\mbf{a} \circ \mbf{b} \circ \mbf{c})_{ijk} = a_ib_jc_k$
is an order-3  tensor and
$(\underline{\mbf  A}  \circ  {\mbf  B})_{i_1,\ldots,i_p,k_1,\ldots,k_q}$
$=a_{i_1,\ldots,i_p}b_{k_1,\ldots,k_q}$
is the outer product of 
$\underline{\mbf A} = \{   a_{i_1,\ldots,i_p} \}$
and
$\underline{\mbf B} = \{ b_{k_1,\ldots,k_q} \}$.

\section{Embedding a model into a vector space}
Let $\{e_1,\ldots,e_N\}$  be the  set of constants  in ${\cal  L}$ and
$\mbf{M}=({\cal D},I)$  a model where ${\cal  D} = \{e_1,\ldots,e_N\}$
(we here identify $I(e_i)$ and $I(r)$  in $\mbf{M}$ with $e_i$ and $r$
respectively  to  avoid notational  complications).   We  show how  to
replace the evaluation $\denote{F}$ of  a prenex formula $F$ in ${\mbf
  M}$ with  the evaluation  of $\Sigma_F$, a  set of  tensors compiled
from  $F$, in  $N$-dimensional  Euclidean  space $\mathds{R}^N$.   The
compilation of $F$ into $\Sigma_F$  starts from literals then proceeds
to compound formulas and quantifications.

\subsection{Entities, literals, logical connectives and existential
 quantifier}
\label{subsec:model}

First  we isomorphically  map  ${\bf  M}$ to  a  model  ${\bf M}'$  in
$\mathds{R}^N$.  We map entities $e_i \in {\cal D}$ to one-hot vectors
${\mbf e}_i$.   So ${\cal D}$  is mapped to
${\cal  D}'= \{{\mbf e}_1,\ldots,{\mbf e}_N\}$, the basis of $\mathds{R}^N$.
We next map a $k$-ary relation $r$ in ${\mbf M}$ to a $k$-ary relation $r'$
over ${\cal  D}'$ which is computed by an  order-$k$ tensor
$\underline{\mbf R}= \{ r_{i_1,\ldots,i_k} \}$.
$\underline{\mbf  R}$ is designed to retain the truth value
$\denote{r(e_{i_1},\ldots,e_{i_k})}$ in {\bf M}
and given by the equation

\begin{eqnarray}
\lefteqn{ \denote{r(e_{i_1},\ldots,e_{i_k})} } \nonumber \\
 & = &  \underline{\mbf R}({\mbf e}_{i_1},\ldots,{\mbf e}_{i_k})
        \quad \mbox{as multi-linear map} \nonumber \\
 & = &  \underline{\mbf R} \times_{1,1}{\mbf e}_{i_1}
              \times_{1,2}\cdots\times_{1,i_k} {\mbf e}_{i_k} \nonumber \\
 & = &  r_{i_1,\ldots,i_k} \in \{1,0\}\;\;
    (\forall i_1,\ldots,i_k \in \{1,\ldots,N\}). \label{tensor:atom}
\end{eqnarray}
We identity   $r'$  with $\underline{\mbf  R}$ for  simplicity
and say $\underline{\mbf  R}$ encodes the {\bf  M}-relation $r$.  Let
${\bf M}'$  be a  model $({\cal D}',I'$)  in $\mathds{R}^N$  such that
$I'$ interprets entities by $I'(e_i)  = {\mbf e}_{i}$ $(1\leq i\leq
N)$ and  relations $r$ by  $I'(r) = \underline{\mbf R}$
introduced by (\ref{tensor:atom}).

We next  inductively define the evaluation  $\denote{F}'_{I',a'}$ of a
formula $F$ in  ${\bf M}'$.  Let $a$ be an  assignment in ${\bf M}$
and $a'$ the corresponding assignment in ${\bf  M}'$, i.e.,  $a(x)= e_i$
if-and-only-if $a'(x) = {\mbf e}_i$.   For   a   ground   atom
$r(e_{i_1},\ldots,e_{i_k})$, we define

\begin{equation}
\denote{r(e_{i_1},\ldots,e_{i_k})}'
  = \underline{\mbf R}({\mbf e}_{i_1},\ldots,{\mbf e}_{i_k})
  \;(\forall i_1,\ldots,i_k \in \{1,\ldots,N\})
\label{tensor:atom2}
\end{equation}
where
$\underline{\mbf R} = \{r_{i_1,\ldots,i_k}\}$ is the tensor
that encodes the {\bf M}-relation $r$ in {\bf M} (see (\ref{tensor:atom})).
By definition $\denote{F}'_{I',a'} = \denote{F}_{I,a}$ holds for any atom $F$.
Negative literals are evaluated specifically in ${\bf M}'$
using tensors $\neg\underline{\mbf R}$ introduced by

\begin{eqnarray}
\denote{\neg r(e_{i_1},\ldots,e_{i_k})}'
& = & \neg\underline{\mbf R}({\mbf e}_{i_1},\ldots,{\mbf e}_{i_k}) 
              \nonumber \\
& = & 1 - r_{i_1,\ldots,i_k} \nonumber \\
\mbox{where}\;\; \neg\underline{\mbf R}
& \defined & \overbrace{\mbf{1}\circ\cdots\circ\mbf{1}}^k
      - \underline{\mbf R}
  \label{tensor:negatom}
\end{eqnarray}
We say $\neg\underline{\mbf R}$ encodes an {\bf M}-relation $\neg r$\footnote{
$\mbf{1}\circ\cdots\circ\mbf{1}$ is an order-$k$ tensor.
$\mbf{1}\circ\cdots\circ\mbf{1}({\mbf e}_{i_1},\ldots,{\mbf e}_{i_k})$
$=(\mbf{1}\bullet {\mbf e}_{i_1})\cdots (\mbf{1}\bullet {\mbf e}_{i_k})
= 1$ holds.
}. Negation other than negative literals and
conjunction and disjunction  are evaluated
in ${\bf M}'$ as follows.

\begin{eqnarray}
\denote{\neg F}'_{I',a'}
& = & 1 - \denote{F}'_{I',a'}    \label{tensor:negation} \\
\denote{F_1 \wedge\cdots\wedge F_h}'_{I',a'}
& = & \denote{F_1}'\cdots \denote{F_h}'_{I',a'}  \label{tensor:and} \\
\denote{F_1 \vee\cdots\vee F_h}'_{I',a'}
& = & {\rm min}_1(\denote{F_1}'_{I',a'}+\cdots+\denote{F_h}'_{I',a'}) \label{tensor:or} \\
\denote{\exists y F}'_{I',a'}
& = & {\rm min}_1
          (\sum_{i=1}^N \denote{{F}_{y\leftarrow e_i}}'_{I',a'})  \label{tensor:exists}
\end{eqnarray}
Here ${\rm min}_1(x) = {\rm min}(x,1) = \mbox{$x$ if $x<1$ else $1$}$
and when applied to tensors, it means componentwise application.
${F}_{y\leftarrow e_i}$ is a formula obtained from  $F$
by replacing every free occurrence of $y$ in $F$ with $e_i$.
Universal quantification is treated as $\forall x F = \neg \exists x \neg F$.

It    is    straightforward    to   check    that    the    evaluation
$\denote{F}'_{I',a'}$   of   a   formula   $F$  in   ${\bf   M}'$   by
(\ref{tensor:atom2}), (\ref{tensor:negatom}), (\ref{tensor:negation}),
(\ref{tensor:and}),   (\ref{tensor:or})    and   (\ref{tensor:exists})
coincides  with $\denote{F}_{I,a}$  in ${\bf  M}$.  However,  although
this evaluation is carried out in a vector space, i.e. $\mathds{R}^N$,
it is based on the reduction  of quantification to the ground level as
(\ref{tensor:exists}) indicates and contains  a lot of redundancy.  We
next show how to do the same thing without grounding quantifications.

\subsection{$\exists$-DNF and $\forall$-CNF as tensors}
Now  we  come  to  the  crucial  point  of  our  proposal,  evaluating
quantified formulas without grounding.  Consider a prenex formula $F =
Q_1x_1\cdots Q_m  x_m G$.   For the  moment we  assume the  inner most
quantified subformula $Q_m x_m G$ is in $\exists$-DNF.

Let $\exists\,y (L_1 \wedge\cdots\wedge L_M)$ be an arbitrary disjunct
of $Q_m x_m G$ where $L_1,\ldots,L_M$ are literals.  We further assume
the {\em variable  condition\/} that $y$ occurs  once in each
literal $L_m = r_m^{\circ}(x^m_1,\ldots,x^m_{N_m})$ ($1\leq m\leq M$).
Here  $r_m^{\circ}  =  r_m$  if  $L_m$  is  a  positive  literal  else
$r_m^{\circ} = \neg r_m$.  $r_m^{\circ}$ in {\bf M} is called the {\bf
  M}-relation contained in $L_m$.
Let $\underline{\mbf R}_m^{\circ}$($1\leq m \leq M$)
be a tensor encoding the {\bf M}-relation $r_m^{\circ}$ defined respectively by 
(\ref{tensor:atom}) or (\ref{tensor:negatom}).
So $\underline{\mbf R}_m^{\circ}(\mbf{x}^m_1,\ldots,\mbf{x}^m_{N_m})
= \denote{L_m[x^m_1,\ldots,x^m_{N_m}]}$ holds where
$x^m_i$ ($1\leq i \leq N_m$)
range over the domain of constants ${\cal  D}= \{ e_1,\ldots,e_N \}$
while $\mbf{x}^m_i$ correspondingly range over the domain of the standard basis
${\cal  D}'= \{ \mbf{e}_1,\ldots,\mbf{e}_N \}$.
The notation $L_m[x^m_1,\ldots,x^m_{N_m}]$ emphasizes that
$x^m_1,\ldots,x^m_{N_m}$ occur in $L_m$.

Suppose $L_m= r_m^{\circ}(x^m_1,\ldots,x^m_{N_m})$ ($1\leq m \leq M$)
has $y$ as the $j_m$-th argument.
Remove $y$, the $j_m$-th argument, from $(x^m_1,\ldots,x^m_{N_m})$.
Write the remaining arguments (with order preserved) collectively as $x^{(m)}_{-y}$ and
consider $ \denote{L_m} = \denote{L_m[x^{(m)}_{-y}]} =
\underline{\mbf R}_m^{\circ}(\mbf{x}^{(m)}_{-y}) $
as a function of $x^{(m)}_{-y}$ 
or a function of the corresponding arguments $\mbf{x}^{(m)}_{-y}$
over ${\cal D}'^{N_m-1}$ parameterized with $y$.
Then consider $\denote{\exists\,y (L_1 \wedge\cdots\wedge L_M)} \in \{1,0\}$
as  a relation combined with arguments (free variables, possibly duplicate)
$(x^{(1)}_{-y},\ldots,x^{(M)}_{-y})$ over ${\cal D}^{\sum_m (N_m-1)}$,
or equivalently, a function applied to $(\mbf{x}^{(1)}_{-y},\ldots,\mbf{x}^{(M)}_{-y})$
over ${\cal D}'^{\sum_m (N_m-1)}$.
We seek a tensor $\underline{\mbf{R}}^{\rm new}$ that encodes this function,
i.e., $\underline{\mbf{R}}^{\rm new}$ such that
$\denote{\exists\,y (L_1 \wedge\cdots\wedge L_M)}
= \underline{\mbf{R}}^{\rm new}(\mbf{x}^{(1)}_{-y},\ldots,\mbf{x}^{(M)}_{-y})$
holds.
Look at

\begin{eqnarray}
\lefteqn{\denote{\exists\,y (L_1 \wedge\cdots\wedge L_M)}} \nonumber\\
& = & \denote{\exists\,y (r_1^\circ(x^{(1)}) \wedge\cdots\wedge r_M^\circ(x^{(M)}))} \nonumber\\
& = &  {\rm min}_1 \Big(
   \sum_{k=1}^N \prod_{m=1}^M
    \underline{\mbf{R}}_m^{\circ}(\mbf{x}^{(m)})_{y\leftarrow \mbf{e}_k}
                     \Big)  \nonumber\\
& = & {\rm min}_1 \Big(
       \sum_{k=1}^N \prod_{m=1}^M (\underline{\mbf{R}}_m^{\circ} \bullet_{j_m} \mbf{e}_k)
            (\mbf{x}^{(m)}_{-y}) \Big) \nonumber\\
& = &  {\rm min}_1 \Big(
        \sum_{k=1}^N 
        \left\{ \big( (\underline{\mbf{R}}_1^{\circ} \bullet_{j_1} \mbf{e}_k)
        \circ\cdots\circ (\underline{\mbf{R}}_M^{\circ} \bullet_{j_M} \mbf{e}_k)  \big)
           (\mbf{x}^{(1)}_{-y},\ldots,\mbf{x}^{(M)}_{-y})
        \right\}   \Big) \nonumber\\
& = &  {\rm min}_1 \Big(
       \sum_{k=1}^N
       \big( (\overbrace{\mbf{e}_k\circ\cdots\circ\mbf{e}_k}^{M})
      \times_{1,j_1} \underline{\mbf{R}}_1^{\circ}
             \times_{1,j_2}\cdots \times_{1,j_M} \underline{\mbf{R}}_M^{\circ}
      \big) \Big) \nonumber\\
&   &  \quad\quad  (\mbf{x}^{(1)}_{-y},\ldots,\mbf{x}^{(M)}_{-y}) \nonumber\\
& = & {\rm min}_1
      \big( \underline{\mbf{Q}}^{\exists,M}
      \times_{1,j_1} \underline{\mbf{R}}_1^{\circ} \times_{1,j_2} \cdots \times_{1,j_M}
                 \underline{\mbf{R}}_M^{\circ}
                 \big) (\mbf{x}^{(1)}_{-y},\ldots,\mbf{x}^{(M)}_{-y}) \nonumber
\end{eqnarray}
Here
\begin{eqnarray}
   \underline{\mbf{Q}}^{\exists,M}  & \defined &
      \sum_{k=1}^N \overbrace{\mbf{e}_k\circ\cdots\circ\mbf{e}_k}^{M}
           \label{quant:exists}
\end{eqnarray}
is a tensor representing the existential quantifier $\exists\,y$.

Summing up, the {\bf M}-relation extracted from $\exists\,y (L_1 \wedge\cdots\wedge L_M)$,
which solely  depends on the free variables in it, is encoded by 
\begin{equation}
\underline{\mbf{R}}^{\rm new}
=
{\rm min}_1
      \big( \underline{\mbf{Q}}^{\exists,M}
      \times_{1,j_1} \underline{\mbf{R}}_1^{\circ} \times_{1,j_2} \cdots \times_{1,j_M}
                 \underline{\mbf{R}}_M^{\circ}
      \big)
\label{tensor:existsand}
\end{equation}
where $\underline{\mbf{R}}^{\circ}_m$ encodes the {\bf M}-relation contained in
$L_m$ ($1\leq m\leq M$) and the existential quantifier $\exists\,y$
that quantifies $M$ free occurrences of $y$
in $L_1 \wedge\cdots\wedge L_M$ 
is encoded by an order-$M$ tensor $\underline{\mbf{Q}}^{\exists,M}$
introduced by (\ref{quant:exists}).
We call the equation (\ref{tensor:existsand})
a definition  for $\underline{\mbf{R}}^{\rm new}$.

Similarly, if $Q_m x_m G$ is
a $\forall$-CNF formula $\forall\,y (L_1 \vee\cdots\vee L_M)$,
the relation in {\bf M} extracted from
$\forall\,y (L_1 \vee\cdots\vee L_M)$ is encoded by
\begin{equation}
\underline{\mbf{R}}^{\rm new}
=
\overbrace{\mbf{1}\circ\cdots\circ\mbf{1}}^{\Sigma_m N_m-1} -
{\rm min}_1
      \big( \underline{\mbf{Q}}^{\exists,M}
      \times_{1,j_1} \underline{\mbf{R}}_1^{\circ} \times_{1,j_2} \cdots \times_{1,j_M}
                 \underline{\mbf{R}}_M^{\circ}
      \big)
\label{tensor:forallor}
\end{equation}
where $\underline{\mbf{R}}^{\circ}_m$ encodes the {\bf M}-relation  contained in
$\neg L_m$ ($1\leq m\leq M$) (details omitted).

\subsection{Compiling prenex formulas}
We now  compile a prenex formula  $F = Q_1x_1\cdots Q_m  x_m G$, using
(\ref{tensor:existsand})   and    (\ref{tensor:forallor}),   into   an
associated  set  $\Sigma_F$  of   tensor  definitions  which  computes
$\denote{F}$ without grounding.  However there is one problem to solve
before compilation; (\ref{tensor:existsand}),  for example, is derived
from $\exists\,y (L_1 \wedge\cdots\wedge  L_M)$ under the the variable
condition.   When  this condition  is  violated,  we need  to  somehow
recover it.

\begin{figure}[hb]
\rule{0.5\textwidth}{0.25mm}\\ [-1em]
\begin{center}
\begin{description}
\item[{\bf Input:}]
A model {\bf M} for a first-order language ${\cal L}$ with finitely many constants
and a first-order closed formula $F = Q_1x_1\cdots Q_m x_m G$ in  ${\cal L}$
in prenex normal form such that no atom has duplicate variables
and G is an open DNF or CNF \\
\mbox{}\\
\item[{\bf Procedure:}]
\item[{\bf [Step 1]}] Set $\Sigma_F = \{ \}$, $G_m = G$ and $F_m = Q_m x_m G_m$;
\item[{\bf [Step 2]}]\mbox{}\\
{\bf For} i = m {\bf down-to} 1 {\bf Do} \\
Write $F_i = Q_i x_i G_i$; \\
If $Q_i = \forall$ then goto {\bf [Step 2-B]}; \\
{\bf [Step 2-A]} \\
Convert $Q_i x_i G_i$ to $\exists$-DNF $G^*_i$; \\
{\bf For each} disjunct $D$ {\bf in} $G^*_i$ {\bf Do} \\
\quad Write $D = \exists\,y (L_1 \wedge\cdots\wedge L_M) \wedge D'$
           where $y$ occurs once \\
\quad\quad in each $L_i$ ($1 \leq i \leq M$) and has no occurrence in $D'$; \\
\quad Let ${x}_{\rm free}$ be an enumeration without duplication of \\
\quad\quad free variables in
        $D'' = \exists\,y (L_1 \wedge\cdots\wedge L_M) $; \\
\quad Define a new atom by $r^{\rm new}({x}_{\rm free}) \Leftrightarrow D''$; \\
\quad Replace $D$ in $G^*_i$ with $r^{\rm new}({x}_{\rm free})\wedge D'$; \\
\quad Introduce a new tensor $\underline{\mbf{R}}^{\rm new}$ 
               by (\ref{tensor:existsand}) encoding \\
\quad\quad the new relation $r^{\rm new}$ in {\bf M}; \\
\quad Add to $\Sigma_F$ the tensor definition for $\underline{\mbf{R}}^{\rm new}$; \\
\quad {\bf endDo} \\
Set $F_{i-1} = Q_{i-1} x_{i-1} G^*_i$; \\
{\bf [Step 2-B]} \\
Convert $Q_i x_i G_i$ to $\forall$-CNF $C^*_i$;\\
\quad (the rest is dual to {\bf [Step 2-A]} and omitted) \\
{\bf endDo} \\
\item[{\bf [Step 3]}]
 If $F_0 = r_1 \wedge\cdots\wedge r_h$
   then put $F^{\rm tensor} = r_1 \cdots r_h$; \\
 Else $F_0 = r_1 \vee\cdots\vee r_h$ and
   put $F^{\rm tensor} = {\rm min}_1(r_1 +\cdots+ r_h)$; \\
\quad ($r_i$'s are atoms with no arguments, equated with \\
\quad\quad  true or false, and hence with $\{1,0\}$) \\
\mbox{}\\
\item[{\bf Output:}]
$F^{\rm tensor}$ with a set $\Sigma_F$ of tensor definitions.
$\Sigma_F$ 
$\Sigma_F$ encodes new {\bf M}-relations appearing in  $S$
and $F_0$ gives $\denote{F}$ in {\bf M}.
\end{description}
\end{center}
\rule{0.5\textwidth}{0.25mm} \\[-1em]
\caption{Compilation procedure of prenex formulas \label{eval:tensor}}
\end{figure}

There are two  cases where the condition is violated.   The first case
is that some atom $r_m(x^{(m)})$  in $G$ has duplicate occurrences of
variables   in   the  arguments   $x^{(m)}$.    In   this  case,   let
$\mbf{R}_m$ be a  tensor encoding the {\bf  M}-relation $r_m$ which
is given  by (\ref{tensor:atom}).  Let $r^{\rm  new}_m(x'^{(m)})$ be a
new   atom  defined   by   $r^{\rm  new}_m(x'^{(m)})   \Leftrightarrow
r_m(x^{(m)})$ where $x'^{(m)}$ is  an enumeration of $x^{(m)}$ without
duplication.   It is  apparent  that a  new  relation $r^{\rm  new}_m$
stands    for    in    {\bf    M}   is    encoded    by    a    tensor
$\underline{\mbf{R}}^{\rm         new}_m$         such         that
$\underline{\mbf{R}}^{\rm        new}_m(\mbf{x}'^{(m)})        =
\underline{\mbf{R}}_m(\mbf{x}^{(m)})$       where      variables
$\mbf{x}'^{(m)}$ and $\mbf{x}^{(m)}$ run over ${\cal D}'$.
We replace every atom in $G$ that violates the variable condition
with a new atom $r^{\rm new}_m(x'^{(m)})$ described above
so that $\underline{\mbf{R}}^{\rm new}_m$
encodes the new {\bf  M}-relation $r^{\rm new}_m$.
Let the result be $G^*$ and consider $F^* = Q_1x_1\cdots Q_m x_m
G^*$.  Obviously  when evaluated in ${\bf  M}'$, $F^*$ and $F$  give the
same result,  i.e., $\denote{F^*}' = \denote{F}'  (=\denote{F})$ holds.
So in the first case, we compile $F^*$ instead of $F$.

The second  case is that, for  example, some $L_i$s in  $D = \exists\,y
(L_1 \wedge\cdots\wedge L_M)$ have no  occurrence of $y$. In this case,
we just shrink  the scope of $\exists\,y$ and rewrite  $D$ like $D =
\exists\,y       (L_1       \wedge\cdots\wedge       L_h)       \wedge
L_{h+1}\wedge\cdots\wedge L_M$.

Taking these modifications into  account, we summarize our compilation
procedure in Figure~\ref{eval:tensor}.   When a model ${\bf  M}$ and a
closed prenex formula $F$ are given, the compilation procedure returns
an algebraic formula  $F^{\rm tensor}$ and a set  $\Sigma_F$ of tensor
definitions.   Evaluating  $F^{\rm  tensor}$  using  $\Sigma_F$  gives
$\denote{F}$, the truth value of $F$ in {\bf M}.

\subsection{A compilation example}

Let $F_{ABCD} = \forall x \exists y ((A(x,y) \wedge B(x)) \vee (C(x,y)\wedge D(y))$.
We compile $F_{ABCD}$ into a set $\Sigma_{F_{ABCD}}$ of tensor definitions
along the compilation procedure in Figure~\ref{eval:tensor}.
Let $\underline{\mbf{A}}$, $\underline{\mbf{B}}$, $\underline{\mbf{C}}$
and $\underline{\mbf{D}}$ respectively be tensors encoding
${\bf M}$-relations $A$, $B$, $C$ and $D$.

Set $\Sigma_{F_{ABCD}} = \{ \}$.
First we convert $F_{ABCD}$'s innermost subformula $F_2$ into $\exists$-DNF:
\begin{eqnarray*}
F_2
 &  = & \exists y ((A(x,y) \wedge B(x)) \vee (C(x,y)\wedge D(y))) \\
 &  = & \exists y (A(x,y) \wedge B(x)) \vee \exists y (C(x,y)\wedge D(y)) \\
 &  = & (\exists y A(x,y) \wedge B(x)) \vee \exists y (C(x,y)\wedge D(y)).
\end{eqnarray*}
Next we introduce new atoms and rewrite $F_2$ to $G^*_2$:
\begin{eqnarray*}
r^{\rm new}_{A}(x) & \Leftrightarrow & \exists y A(x,y) \\
r^{\rm new}_{CD}(x) & \Leftrightarrow & \exists y (C(x,y)\wedge D(y)) \\
G^*_2 & = &  (r^{\rm new}_{A}(x) \wedge B(x)) \vee r^{\rm new}_{CD}(x).
\end{eqnarray*}
Correspondingly to these new atoms, we construct tensors below
which encode the corresponding relations in {\bf M}
and add them to $\Sigma_F$:
\begin{eqnarray}
\underline{\mbf{R}}^{\rm new}_{A}
& = &  {\rm min}_1( \underline{\mbf{Q}}^{\exists,1} \times_{1,2}
          \underline{\mbf{A}})
             \label{newten:A} \\
\underline{\mbf{R}}^{\rm new}_{CD}
& = &  {\rm min}_1( \underline{\mbf{Q}}^{\exists,2} \times_{1,2}
          \underline{\mbf{C}} \times_{1,1} \underline{\mbf{D}}).
             \label{newten:CD}
\end{eqnarray}

We put $F_1 = \forall x F_2 = \forall x G^*_2$ and continue compilation.
We convert $F_1$ to $\forall$-CNF:
\begin{eqnarray*}
F_1 & = & \forall x ((r^{\rm new}_{A}(x) \wedge B(x)) \vee r^{\rm new}_{CD}(x)) \\
    & = & \forall x (r^{\rm new}_{A}(x)\vee r^{\rm new}_{CD}(x)) \wedge
           \forall x (B(x) \vee r^{\rm new}_{CD}(x)).
\end{eqnarray*}
We introduce new atoms and rewrite $F_1$ to $G^*_1$:
\begin{eqnarray*}
r^{\rm new}_{ACD} & \Leftrightarrow & \forall x\, r^{\rm new}_{A}(x) \vee r^{\rm new}_{CD}(x) \\
r^{\rm new}_{BCD} & \Leftrightarrow & \forall x\, r_{B}(x) \vee r^{\rm new}_{CD}(x) \\
G^*_1 & = & r^{\rm new}_{ACD} \wedge r^{\rm new}_{BCD}.
\end{eqnarray*}
We construct tensors (scalars) for $r^{\rm new}_{ACD}$ and $r^{\rm new}_{BCD}$:
\begin{eqnarray}
\underline{\mbf{R}}^{\rm new}_{ACD}
 & = & 1 - {\rm min}_1( \underline{\mbf{Q}}^{\exists,2} \times_{1,1}
       \neg\underline{\mbf{R}}^{\rm new}_{A}  \times_{1,1}
       \neg\underline{\mbf{R}}^{\rm new}_{CD})
             \label{newten:ACD} \\
\underline{\mbf{R}}^{\rm new}_{BCD}
 & = & 1 - {\rm min}_1 ( \underline{\mbf{Q}}^{\exists,2} \times_{1,1}
       \neg\underline{\mbf{B}} \times_{1,1}
       \neg\underline{\mbf{R}}^{\rm new}_{CD} )
             \label{newten:BCD}
\end{eqnarray}
and add them to $\Sigma_{F_{ABCD}}$.
Now $\Sigma_{F_{ABCD}} = \{(\ref{newten:A}), (\ref{newten:CD}),
(\ref{newten:ACD}),(\ref{newten:BCD}) \}$.
Finally we put
\begin{eqnarray*}
F_0 & = & G^*_1 \;=\; r^{\rm new}_{ACD} \wedge r^{\rm new}_{BCD}. \\
F_{ABCD}^{\rm tensor}
 & = & \underline{\mbf{R}}^{\rm new}_{ACD} \cdot
       \underline{\mbf{R}}^{\rm new}_{BCD}.
\end{eqnarray*}

So $\denote{F_{ABCD}}$ in  {\bf M} is evaluated  without grounding by
computing   $F_{ABCD}^{\rm   tensor}$   using   $\Sigma_{F_{ABCD}}   =
\{(\ref{newten:A}),                                 (\ref{newten:CD}),
(\ref{newten:ACD}),(\ref{newten:BCD}) \}$.

\subsection{Binary predicates: matrix compilation}

The compilation procedure in  Figure~\ref{eval:tensor} is general. It
works  for arbitrary  prenex formulas  $F$ with  arbitrary predicates.
However  when  $r$ is  a  binary  predicate, the  corresponding  tensor
$\underline{\mbf  R}$  is a  bilinear  map  and represented  by  an
adjacency matrix ${\mbf R}$ as follows.

\begin{eqnarray}
\denote{r(e_i,e_j)}
& = & ({\mbf e}_i \bullet {\mbf R}{\mbf e}_j)
    \;=\; {\mbf e}_i^T {\mbf R}{\mbf e}_j
    \;=\; r_{ij}  \in \{1,0\}  \label{tensor:biatom}
\end{eqnarray}

In  such  binary  cases,  we can  often  ``optimize''  compilation  by
directly compiling $F$ using  matrices without introducing $\Sigma_F$.
This is quite important in processing  KGs logically as they are a set
of ground  atoms with  binary predicates.  Hence  we here  derive some
useful    compilation   patterns    using    matrices   defined    by
(\ref{tensor:biatom}) for  formulas with binary predicates.   We
specifically adopt
$\denote{F}_{\bf Mat}$ to denote the result of compilation
using matrices that faithfully follows
(\ref{tensor:atom2}), (\ref{tensor:negatom}), (\ref{tensor:negation}),
(\ref{tensor:and}),   (\ref{tensor:or})    and   (\ref{tensor:exists})
in Subsection~\ref{subsec:model}.

\begin{eqnarray}
\lefteqn{
  \denote{\exists y\, r_1(x,y)\wedge r_2(y,z)}_{\bf Mat} }  \nonumber\\
  & = & \denote{ (r_1(x,e_1)\wedge r_2(e_1,z))\vee\dots\vee
              (r_1(x,e_N)\wedge r_2(e_N,z)) }_{\bf Mat} \nonumber\\
  & = & {\rm min}_1\Big(\sum_{j=1}^N 
               \denote{ r_1(x,e_j)\wedge r_2(e_j,z) }_{\bf Mat}\Big) \nonumber\\
  & = & {\rm min}_1\Big( \sum_{j=1}^N
           {{\mbf x}}^T{\mbf R}_1{\mbf e}_{j}
           {{\mbf e}_{j}}^T{\mbf R}_2{\mbf z} \Big) \nonumber\\
  & = & {\mbf x}^T
           {\rm min}_1\Big( {\mbf R}_1
          \big( \sum_{j=1}^N{\mbf e}_{j}{{\mbf e}_{j}}^T \big){\mbf R}_2 \Big)
             {\mbf z} \nonumber \\
  & = &  {{\mbf x}}^T
            {\rm min}_1\big( {\mbf R}_1{\mbf R}_2 \big){\mbf z}  \label{matrix:ex24}
\end{eqnarray}
Here ${\mbf x}$ and ${\mbf z}$ run over
${\cal D}'= \{ {\mbf e}_1,\ldots, {\mbf e}_N \}$.
Hence the synthesized relation $r_{12}(x,y) \defined \exists y\, r_1(x,y)\wedge r_2(y,z)$
is encoded by a matrix ${\mbf R}_{12} = {\rm min}_1\big( {\mbf R}_1{\mbf R}_2 \big)$.
What is important with this example, or with binary predicates in general, is the fact that
$\mbf{Q}^{\exists,2} = \sum_{j=1}^N{\mbf e}_{j}{{\mbf e}_{j}}^T = I$,
an identity matrix, holds.

Similarly by applying (\ref{matrix:ex24}), we can compile
a doubly quantified formula $\exists x\,\exists y\, r_1(x,y)\wedge r_2(x,y)$
as follows\footnote{
When  $r(x,y)$ is encoded by $\mbf{R}$ as $(\mbf{x} \bullet \mbf{R}\mbf{y})$,
$r(y,x)$ is encoded by $\mbf{R}^T$ because
$(\mbf{y} \bullet \mbf{R}\mbf{x}) =  (\mbf{x} \bullet \mbf{R}^T\mbf{y})$
holds.
}.

\begin{eqnarray}
\lefteqn{
  \denote{\exists x\,\exists y\, r_1(x,y)\wedge r_2(x,y)}_{\bf Mat} } \nonumber\\
  & = & \denote{ \left(\exists y\,r_1(e_{1},y)\wedge r_2(e_1,y)\right)\vee\dots\vee
           \left(\exists y\,r_1(e_{N},y)\wedge r_2(e_N,y)\right) }_{\bf Mat} \nonumber\\
  & = & {\rm min}_1 \big( \sum_{i=1}^N
           {{\mbf e}_{i}}^T
               {\rm min}_1( {\mbf R}_1{\mbf R}_2^T ) {\mbf e}_{i} \big)  \nonumber\\
  & = &  {\rm min}_1({\rm tr}({\mbf R}_1{\mbf R}_2^T)) \label{matrix:ex2}
\end{eqnarray}

Hence, a Horn formula $\forall x\,\forall y\, r_1(x,y)\Rightarrow r_2(x,y)$
is compiled into

\begin{eqnarray}
\lefteqn{ \denote{\forall x\,\forall y\, r_1(x,y)\Rightarrow r_2(x,y)}_{\bf Mat} } \\
  & = & \denote{ \neg\, \exists x\,\exists y\, r_1(x,y)\wedge \neg r_2(x,y) }_{\bf Mat} \nonumber\\
  & = & 1 - {\rm min}_1 ({\rm tr}({\mbf R}_1{\neg{\mbf R}_2}^T)).  \label{matrix:horn1}
\end{eqnarray}
Note that ${\rm tr}({\mbf R}_1{\neg{\mbf R}_2}^T)$ gives
the number of pairs $(x,y)$ that do not satisfy $r_1(x,y)\Rightarrow r_2(x,y)$.
Consequently ${\rm tr}({\mbf R}_1{\neg{\mbf R}_2}^T)=0$
implies every pair $(x,y)$ satisfies $r_1(x,y)\Rightarrow r_2(x,y)$
and vice versa. Our compilation is thus confirmed correct.

Another, typical, Horn formula $\exists\,y\,r_1(x,y)\wedge r_2(y,z)\Rightarrow r_3(x,z)$
is compiled into

\begin{eqnarray}
\lefteqn{
  \denote{ \forall x\,\forall z\,
      \big( \exists\,y\,r_1(x,y)\wedge r_2(y,z)\Rightarrow r_3(x,z) \big) }_{\bf Mat} } \nonumber \\
  & = & 1 - {\rm min}_1\big(
      {\rm tr}( {\rm min}_1({\mbf R}_1{{\mbf R}_2}){\neg{\mbf R}_3}^T )
                      \big). \label{matrix:horn2}
\end{eqnarray}
Again ${\rm tr}\big({\rm min}_1( {\mbf R}_1{{\mbf R}_2}){\neg{\mbf R}_3}^T \big)$
is the total  number of $(x,z)$s that do not satisfy
$\exists\,y\, r_1(x,y)\wedge r_2(y,z)\Rightarrow r_3(x,z)$.
So our compilation is correct.

\subsection{Recursive matrix equations}

Our  non-grounding linear-algebraic  approach yields  tensor equations
from logical equivalence, and this property provides a new approach to
the  evaluation of  Datalog  programs.   We sketch  it  using a  small
example.   Consider the  following Datalog  program that  computes the
transitive closure {\tt r2} of a binary relation {\tt r1}.\\

\begin{tabular}{l}
\qquad {\tt r2(X,Z):- r1(X,Z).} \\
\qquad {\tt r2(X,Z):- r1(X,Y),r2(Y,Z).}
\end{tabular}\\

\noindent
This program  defines the least  Hearbrand model ${\bf M}$  where {\tt
  r1} is interpreted  as $r_1$ and {\tt r2}  as $r_1$.  $r_2(x_1,x_h)$
holds  true if-and-only-if  there is  a chain  $x_1,x_2,\ldots,x_h \in
{\bf        M}$       ($        h\geq       1$)        such       that
$r_1(x_1,x_2),r_1(x_2,x_3),\ldots,r_1(x_{h-1},x_h)$  are  all true  in
{\bf M}. Then we see the logical equivalence

\begin{eqnarray}
r_2(x,z) & \Leftrightarrow & r_1(x,z) \vee \exists y ( r_1(x,y) \wedge r_2(y,z) )
    \label{trcl:equiv}
\end{eqnarray}

holds for all $x,z$  in {\bf M}. That means
\begin{eqnarray}
\denote{r_2(x,z)}
 & = & \denote{ r_1(x,z) \vee \exists y ( r_1(x,y) \wedge r_2(y,z) ) }
    \label{trcl:semeq}
\end{eqnarray}
also holds for any $x,z$. Let $\mbf{R}_1$ and $\mbf{R}_2$ be adjacency
matrices  encoding  $r_1$  and  $r_2$ in  {\bf  M}  respectively.   We
translate (\ref{trcl:semeq})  in terms of $\mbf{R}_1$  and $\mbf{R}_2$
as follows.
\begin{eqnarray*}
\mbf{x}^T\mbf{R}_2 \mbf{z}
  & = & \denote{ r_2(x,z) } \\
  & = & \denote{ r_1(x,z) \vee \exists y ( r_1(x,y) \wedge r_2(y,z) } \\
  & = & {\rm min}_1(\mbf{x}^T\mbf{R}_1 \mbf{z}
         + \mbf{x}^T{\rm min}_1(\mbf{R}_1\mbf{R}_2) \mbf{z}) \\
  & = & \mbf{x}^T{\rm min}_1(\mbf{R}_1 + \mbf{R}_1\mbf{R}_2) \mbf{z}
\end{eqnarray*}
Since $\mbf{x}, \mbf{z} \in {\cal D}'$ are arbitrary,
we reach a recursive equation
\begin{eqnarray}
\mbf{R}_2 & = & {\rm min}_1(\mbf{R}_1 + \mbf{R}_1\mbf{R}_2).
  \label{trcl:mateq}
\end{eqnarray}

It is to be noted that when considered an equation for
unknown $\mbf{R}_2$, (\ref{trcl:mateq}) may have more than
one solution\footnote{
For example, $\mbf{R}_2 = \mbf{1} \circ \mbf{1}$ is a solution.
} but we can prove that the transitive closure is the ``least'' solution
of (\ref{trcl:mateq}) in  the sense of matrix ordering\footnote{
Matrices $\mbf{A}=\{ a_{ij} \}$ and $\mbf{B}=\{ b_{ij} \}$ are ordered
by  $\mbf{A}  \leq  \mbf{B}$: $\mbf{A}  \leq  \mbf{B}$  if-and-only-if
$a_{ij} \leq b_{ij}$ for all $i,j$.
}(proof omitted).

Since (\ref{trcl:mateq}) is a nonlinear  equation due to ${\rm min}_1$
operation, it  looks impossible  to apply a  matrix inverse  to obtain
$\mbf{R}_2$.  However we found a way to circumvent this difficulty and
proved  that  it  is  possible  to  obtain  $\mbf{R}_2$  by  computing
(\ref{trcl:mateq_sol_1}) and (\ref{trcl:mateq_sol_2}) as follows.

\begin{eqnarray}
\mbf{R}_2 & = & (\mbf{R}_2^{\dagger})>0  \label{trcl:mateq_sol_1} \\
\mbf{R}_2^{\dagger}
         & = &
    (\mbf{I}-\epsilon\mbf{R}_1)^{-1}   \epsilon\mbf{R}_1   \label{trcl:mateq_sol_2} \\
& & {\rm where}\; \epsilon   = (1+\| \mbf{R}_1 \|_{\infty})^{-1} \nonumber
\end{eqnarray}

\noindent
Here  $(\mbf{R}_2^{\dagger})>0$ means  to  threshold  all elements  in
$\mbf{R}_2$  at  $0$,  i.e,  positive  ones are  set  to  1,  o.w.  to
0\footnote{
The proof and details are stated in an accompanying paper submitted
for publication.
}.

\section{Experiment with transitive closure computation}

We compared our  linear algebraic approach to  Datalog evaluation with
state-of-the-art symbolic  approaches using two tabled  Prolog systems
(B-Prolog \cite{Zhou10}  and XSB  \cite{Swift12}) and two  ASP systems
(DLV  \cite{Alviano10}  and   Clingo  \cite{Gebser14}).   Although  we
conducted  a number  of  experiments computing  various programs  with
artificial and  real data, due to  space limitations, we here  pick up
one example that  computes the transitive closure  of random matrices.
In the experiment\footnote{
All  experiments  are carried  out  on  a  PC with  Intel(R)  Core(TM)
i7-3770@3.40GHz CPU, 28GB memory.
}, we generate  random adjacency matrices by specifying  the number of
dimension $N$  and the  probability $p_e$  of each  entry being  1 and
compute      their     transitive      closure     matrices      using
(\ref{trcl:mateq_sol_1}) and  (\ref{trcl:mateq_sol_2}).
We set $N=1000$  and vary $p_e$ from $0.0001$
to  $1.0$ and  measure the  average  computation time  over five  runs
(details omitted).

\begin{table}[h]
{\small
\begin{tabular}{rrrrrr}\\\hline
 $p_e$    &  Matrix  & B-Prolog &    XSB  &   DLV   &   Clingo  \\ \hline\hline
0.0001    &  0.096   &   0.000  &   0.000 &   0.000 &     0.000 \\
0.001     &  0.094   &   0.004  &   0.003 &   0.293 &     0.038 \\
0.01      &  0.117   &   2.520  &   1.746 &  10.657 &    14.618 \\
0.1       &  0.105   &  18.382  &  16.296 &  75.544 &   125.993 \\
1.0       &  0.100   & 188.280  & 137.903 & 483.380 & 1,073.301 \\ \hline
\end{tabular}
}
\caption{Average computation time for transitive closure computation (sec)}
\label{tab:trcl}
\end{table}

Table~\ref{tab:trcl}  shows  the  result.    Our  approach  is  termed
``Matrix''  in the  table.   Two observations  are  clear.  First  the
computation time  of our  approach, Matrix,  is almost  constant while
others seem linear w.r.t. $p_e$.  Second,  when $p_e$ is small, $p_e =
0.0001 \sim  0.001$ and matrices  are sparse, the Matrix  method takes
more time  than existing systems but  when $p_e$ gets bigger,  it runs
orders of  magnitude faster  than them. The  same observation  is made
with other programs (details omitted).

\section{Related work}
There is not much literature  concerning first-order logic embedded in
vector spaces.   The most related work  to ours is a  formalization of
first-order        logic       in        tensor       spaces        by
Grefenstette\cite{Grefenstette13}.     He   actually    proposed   two
formalizations. The first one  represents entities by one-hot vectors,
predicates by  adjacency tensors  and truth values  by two-dimensional
vectors (true by $\top = [1,0]^T$  false by $\bot = [0,1]^T$). AND and
OR are order-3 order tensors whereas NOT is a $2 \times 2$ matrix that
maps $\top$  to $\bot$  and vice versa.   The first  formalization can
completely formalize  a quantifier-free fragment of  first-order logic
in finite domains.   The second formalization represents  a finite set
by a  vector of  multiple ones  (and zeros) and  can deal  with single
quantification by  $\exists$ and $\forall$, but  nested quantification
is  out   of  scope.   The   unification  of  the  first   and  second
formalizations remains an open problem to his tensor approach.

%
%
Krompass  et   al.~\cite{Krompass14}  proposed  a  way   of  answering
existential queries  of the form $\exists  x\, Q_1 \wedge Q_2$  in the
context  of low-dimensional  embeddings.
%
%
Their  approach however does
not assign  an independent representation to  existential quantifiers
and is limited to  a narrow class of the form  $\exists x\, Q_1 \wedge
Q_2$.

We  found  no literature  on  computing  the  least model  of  Datalog
programs  via solving  recursive matrix  equations.  So  the transitive
closure  computation presented  in this  paper is  possibly the  first
example of this kind.

\section{Conclusion}
We proposed a general approach to evaluate first-order formulas $F$ in
prenex normal form in vector spaces. Given a finite model {\bf M} with
$N$ entities,  we compile  $F$ into a  set $\Sigma_F$  of hierarchical
tensor definitions (equations) with  a nonlinear operation.  Computing
$\Sigma_F$ in  $\mathds{R}^N$ yields  the truth value  $\denote{F}$ in
{\bf  M}.    In  this   compilation  process,   tensor  representation
$\underline{\mbf{Q}}^{\exists,M}$   is   introduced   to   existential
quantifiers themselves  for the first time  as far as we  know.  Since
our  approach does  not  rely on  propositionalization of  first-order
formulas, it  can derive  tensor equations from  logical equivalences.
We exploited  this property  to derive  recursive matrix  equations to
evaluated   Datalog   programs.   We  empirically   demonstrated   the
effectiveness of our linear algebraic approach by showing that it runs
orders  of magnitude  faster  than existing  symbolic approaches  when
matrices are not too sparse.


\begin{thebibliography}{}

\bibitem[\protect\citeauthoryear{Alviano \bgroup et al\mbox.\egroup
  }{2010}]{Alviano10}
Alviano, M.; Faber, W.; Leone, N.; Perri, S.; Pfeifer, G.; and Terracina, G.
\newblock 2010.
\newblock {The Disjunctive Datalog System DLV}.
\newblock In de~Moor, O.; Gottlob, G.; Furche, T.; and Sellers, A., eds., {\em
  Datalog Reloaded, LNCS 6702}. Springer.
\newblock  282--301.

\bibitem[\protect\citeauthoryear{Bollacker \bgroup et al\mbox.\egroup
  }{2008}]{Bollacker08}
Bollacker, K.; Evans, C.; Paritosh, P.; Sturge, T.; and Taylor, J.
\newblock 2008.
\newblock {Freebase: a collaboratively created graph database for structuring
  human knowledge}.
\newblock In {\em Proceedings of the 2008 ACM SIGMOD International Conference
  on Management of data},  1247--1250.

\bibitem[\protect\citeauthoryear{Cichocki \bgroup et al\mbox.\egroup
  }{2009}]{Cichocki09}
Cichocki, A.; Zdunek, R.; Phan, A.-H.; and Amari, S.
\newblock 2009.
\newblock {\em Nonnegative Matrix and Tensor Factorizations: Applications to
  Exploratory Multi-way Data Analysis and Blind Source Separation}.
\newblock John Wiley \& Sons, Ltd.

\bibitem[\protect\citeauthoryear{Gebser \bgroup et al\mbox.\egroup
  }{2014}]{Gebser14}
Gebser, M.; Kaminski, R.; Kaufmann, B.; and Schaub, T.
\newblock 2014.
\newblock {Clingo = ASP + Control: Preliminary Report}.
\newblock volume arXiv:1405.3694v1,  1--9.

\bibitem[\protect\citeauthoryear{Grefenstette}{2013}]{Grefenstette13}
Grefenstette, E.
\newblock 2013.
\newblock {Towards a Formal Distributional Semantics: Simulating Logical
  Calculi with Tensors}.
\newblock {\em Proceedings of the Second Joint Conference on Lexical and
  Computational Semantics}  1--10.

\bibitem[\protect\citeauthoryear{Guu, Miller, and Liang}{2015}]{Guu15}
Guu, K.; Miller, J.; and Liang, P.
\newblock 2015.
\newblock {Traversing Knowledge Graphs in Vector Space}.
\newblock In {\em Proceedings of the 2015 Empirical Methods in Natural Language
  Processing (EMNLP)},  318--327.

\bibitem[\protect\citeauthoryear{Kolda and Bader}{2009}]{Kolda09}
Kolda, T.~G., and Bader, B.~W.
\newblock 2009.
\newblock Tensor decompositions and applications.
\newblock {\em SIAM REVIEW} 51(3):455--500.

\bibitem[\protect\citeauthoryear{Krompa{\ss}, Nickel, and
  Tresp}{2014}]{Krompass14}
Krompa{\ss}, D.; Nickel, M.; and Tresp, V.
\newblock 2014.
\newblock {Querying Factorized Probabilistic Triple Databases}.
\newblock In {\em Proceedings of the 13th International Semantic Web
  Conference({ISWC}'14)},  114--129.

\bibitem[\protect\citeauthoryear{Nickel \bgroup et al\mbox.\egroup
  }{2015}]{Nickel15}
Nickel, M.; Murphy, K.; Tresp, V.; and Gabrilovich, E.
\newblock 2015.
\newblock {A Review of Relational Machine Learning for Knowledge Graphs: From
  Multi-Relational Link Prediction to Automated Knowledge Graph Construction}.
\newblock {\em CoRR} abs/1503.00759.

\bibitem[\protect\citeauthoryear{Rockt{\"a}schel, Singh, and
  Riedel}{2015}]{Rocktaschel15b}
Rockt{\"a}schel, T.; Singh, S.; and Riedel, S.
\newblock 2015.
\newblock {Injecting Logical Background Knowledge into Embeddings for Relation
  Extraction}.
\newblock In {\em Annual Conference of the North American Chapter of the
  Association for Computational Linguistics (NAACL)}.

\bibitem[\protect\citeauthoryear{Swift and Warren}{2012}]{Swift12}
Swift, T., and Warren, D.
\newblock 2012.
\newblock {XSB}: Extending prolog with tabled logic programming.
\newblock {\em Theory and Practice of Logic Programming (TPLP)}
  12(1-2):157--187.

\bibitem[\protect\citeauthoryear{Yang \bgroup et al\mbox.\egroup
  }{2015}]{Yang15}
Yang, B.; Yih, W.; He, X.; Gao, J.; and Deng, L.
\newblock 2015.
\newblock {Embedding Entities and Relations for Learning and Inference in
  Knowledge Bases}.
\newblock In {\em Proceedings of the International Conference on Learning
  Representations ({ICLR}) 2015}.

\bibitem[\protect\citeauthoryear{Zhou, Kameya, and Sato}{2010}]{Zhou10}
Zhou, N.-F.; Kameya, Y.; and Sato, T.
\newblock 2010.
\newblock Mode-directed tabling for dynamic programming, machine learning, and
  constraint solving.
\newblock In {\em {Proceedings of the 22th International Conference on Tools
  with Artificial Intelligence (OCTAL-2010)}},  213--218.

\end{thebibliography}

\end{document}